\documentclass{article}

\PassOptionsToPackage{numbers, compress}{natbib}

\usepackage[final]{nips_2018}

\usepackage[utf8]{inputenc} 
\usepackage[T1]{fontenc}    
\usepackage{hyperref}       
\usepackage{url}            
\usepackage{booktabs}       
\usepackage{amsfonts}       
\usepackage{nicefrac}       
\usepackage{microtype}      
\usepackage{amsmath}
\usepackage{todonotes}
\usepackage{xcolor}
\usepackage{tikz}
\usepackage{mathdots}
\usepackage{yhmath}
\usepackage{cancel}
\usepackage{color}
\usepackage{siunitx}
\usepackage{array}
\usepackage{multirow}
\usepackage{amssymb}
\usepackage{gensymb}
\usepackage{enumitem}
\usepackage{graphicx}
\usepackage[position=top]{subfig}
\usepackage[normalem]{ulem}
\usepackage[font=small,labelfont=bf]{caption}
\usepackage{algorithm}
\usepackage[noend]{algpseudocode}

\newcommand{\xhdr}[1]{{\noindent\bfseries #1}.}

\newcommand{\cut}[1]{}

\title{Compositional Language Understanding with Text-based Relational Reasoning}

\author{
  Koustuv Sinha\thanks{Work done while being an intern at Samsung Advanced Institute of Technology (SAIT), Montreal} \\
  Mila \\
  McGill University, Canada\\
  \texttt{koustuv.sinha@mail.mcgill.ca} \\
  \And
  Shagun Sodhani \\
  Mila\\
  Universit\'{e} de Montr\'{e}al, Canada\\
  \texttt{sshagunsodhani@gmail.com} \\
  \And
  William L. Hamilton \\
  Mila \\
  McGill University, Canada \\
  Facebook AI Research (FAIR), Montreal \\
  \texttt{will.leif.hamilton@gmail.com } \\
  \And
  Joelle Pineau \\
  Mila \\
  McGill University, Canada \\
  Facebook AI Research (FAIR), Montreal \\
  \texttt{jpineau@cs.mcgill.ca} \\
}

\begin{document}

\maketitle

\begin{abstract}
  Neural networks for natural language reasoning have largely focused on extractive, fact-based question-answering (QA) and common-sense inference. 
  However, it is also crucial to understand the extent to which neural networks can perform relational reasoning and combinatorial generalization from natural language---abilities that are often obscured by annotation artifacts and the dominance of language modeling in standard QA benchmarks. 
  In this work, we present a novel benchmark dataset for language understanding that isolates performance on relational reasoning. 
  We also present a neural message-passing baseline and show that this model, which incorporates a relational inductive bias, is superior at combinatorial generalization compared to a traditional recurrent neural network approach. 
\end{abstract}

\section{Introduction}
Neural language understanding systems have been extremely successful at information extraction tasks, such as question answering (QA). 
An array of existing datasets are available, which test a system's ability to extract factual answers text \citep{Rajpurkar2016-yc, Nguyen2016-ec, Trischler2016-fc, Mostafazadeh2016-hu, Su2016-so}, as well as datasets that emphasize simple, commonsense inference (e.g., entailment between sentences) \citep{bowman2015large, williams2018broad}. 
However, it is difficult to evaluate a model's reasoning ability in isolation using existing datasets. 
Most datasets combine several challenges of language processing into one, such as co-reference / entity resolution, incorporating world knowledge, and semantic parsing.
Moreover, the state-of-the-art on all these existing benchmarks relies heavily on large, pre-trained language models \cite{devlin2018bert,peters2018deep}, highlighting that the primary difficulty in these datasets is incorporating the statistics of natural language, rather than reasoning. 


In this work, we see to directly evaluate and innovate on the compositional reasoning ability of a QA system. Inspired by CLEVR \citep{Johnson2016-mw}---a synthetic computer vision dataset that isolates the challenges of relational reasoning---we propose a text based dataset for Compositional Language Understanding with Text-based Relational Reasoning (CLUTRR). 
Our initial version, CLUTTR v0.1, requires reasoning and generalizing about kinship relationships, and we plan to use our proposed data generation pipeline to extend the set of tasks in the future. 
We develop and evaluate strong baselines on CLUTTR v0.1, including a recurrent LSTM model and a message-passing graph neural network (GNN).
Our results highlight that the GNN, which incorporates a strong relational inductive bias, outperforms the LSTM at tasks requiring combinatorial generalization. 


\section{The CLUTRR dataset}

To move away from fact-based extractive Q\&A towards more relational reasoning, we consider the classic game of deducing family relations from text. Family relations have been used extensively in automated Knowledge Base (KB) completion tasks \cite{Kok2007-nb,Muggleton1991-dh,Lavrac1994-pr,Rocktaschel2017-ho}.
However, all of these systems operate on (sub)symbolic representations and partial KBs, i.e, they represent an entity \textit{and} a relation with predicate symbols and are provided with a partial KBs. 
In this work, we instead learn directly from the textual descriptions, and we perform Cloze-style anonymization of entities.
 Thus, in our setting the model is not provided with a KB, and, instead, it must learn to \textit{create} an (implicit) knowledge base as it deems fit.
This setup is inspired both the CLEVR task \cite{Johnson2016-mw} as well as the BAbI  tasks \citep{Weston2015-is}.
Like BAbI, we emphasize reasoning in a controlled setting.
However, going beyond the BAbI tasks, we introduce large amounts of distractor information, longer reasoning chains, and the primary focus of our dataset is that it explicitly requires combinatorial generalization, e.g., by training models on examples with $k$ supporting facts and testing on examples that require ${>k}$ supporting facts.
\begin{figure}[t]
\centering
\resizebox{\textwidth}{!}{\input{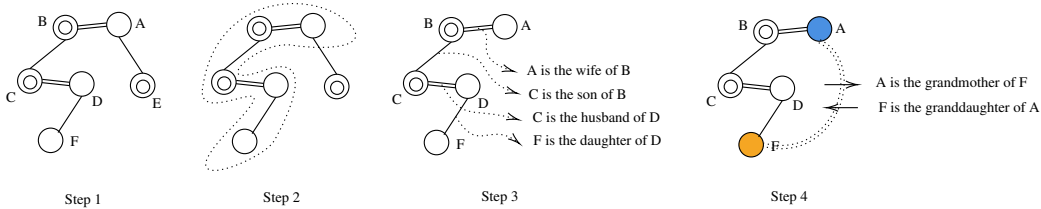}}
\caption{Data generation pipeline. Step 1: generate a kinship graph. Step : sample a path. Step 3: generate story by describing individual edges. Step 4: Predict relation between the first and the last nodes of the path.}\label{fig:data}
\vspace{-10pt}
\end{figure}

\xhdr{Dataset construction}
\label{subsec:data_const}
The core idea behind the CLUTRR task is the following: given a text-based story describing a subset of a kinship graph, the goal is to predict the relationship between two entities, whose relationship is {\em not} stated in the story.
Figure \ref{fig:data} illustrates the data generation process. 
Each example story, $S$, is a sequence of simple sentences, constructed in 3 steps:
\begin{enumerate}[leftmargin=15pt, topsep=0pt, parsep=0pt]
    \item 
    Generate a random kinship graph $G$ with $m$ nodes. The nodes in the graph are the members of the \textit{family} being represented and the edges represents relationships between the members.  The relations belong to the  relation set $R = $ \{\textit{father, mother, son, daughter, husband, wife, grandfather, grandmother, grandson, granddaughter, brother, sister, father-in-law, mother-in-law, son-in-law}\}.
    \item 
    Sample a simple path of $k$ edges from this family graph. 
    \item
    For each edge in the sampled path, a template-based natural language description of the relation is sampled (e.g., ``A is the wife of B'') and added to the story (with 2-3 possible templates per relation).    Additionally, for each of the $k+1$ nodes in the path, 8 ``distractor'' sentences are sampled based on a set of distractor attributes (e.g., ``A works at Y'') and randomly inserted in the story. 
\end{enumerate}
 Given a generated story, $S$, the goal is to predict the relationship between the first entity $e_0$ and the last entity $e_{k}$ in the sampled path that was used to generate the story. 
Thus, the final task is $|R|$-way supervised classification, and to predict the correct relation, a model must infer the implicit knowledge graph represented by the story, reason about missing relationships in this implicit knowledge graph, and learn about logical regularities in kinship graphs (e.g., the child of a child is a grandchild). 
The Appendix contains further details on the graph generation pseudocode, natural language templates, distractor sentences, and sampling procedures.


\xhdr{CLUTRR v0.1}
\label{subsec:cluttr}
Our framework is highly extensible and modular, and in this work we focus on a prototype version, termed CLUTRR v0.1\footnote{\url{https://github.com/koustuvsinha/clutrr}}, where we use the above methodology to sample 5000 stories for each possible path length $k$, ranging from $k=3$ to $k=6$.
We denote the sampled subsets for different paths lengths as $M_1, ..., M_6$, and for each of these subsets we use 4000 examples for training and 1000 for testing.
Separating the dataset into subsets based on the length of the underlying reasoning path allows us to explicitly evaluate how well models can generalize (e.g., training on $M_k$ and testing on $M_{k+1}$). 
Future versions will add realistic natural language variation via crowdsourcing. 

\section{Model}

We introduce two strong baseline models for the CLUTRR dataset: the first model is based on a LSTM recurrent neural network and does not incorporate any relational inductive bias. The second model is a novel graph neural network (GNN) \cite{Gilmer2017} architecture that is specifically designed to capture the relational structure of the CLUTRR reasoning task.


\subsection{Setup}

The input to the model is a story  $S = (s_1, s_2, .., s_n)$, where $s_i = (w^i_1, w^i_2, ..., w^i_m)$ is a sentence consisting of $m$ words, each represented by a $d$-dimensional embedding. 
A subset of these words are the entities $\{e_1, e_2, ..., e_k\}$ (with Cloze-style anonymization), where each entity denotes a node in the latent kinship graph.  Each sentence describes a relation between a pair of entities, or a ``distractor'' attribute of one of the entities. The task is therefore to use the input sentence to predict the relation between a pair of entities $(e_i, e_j)$ as a $|R|$-way classification problem (see Section \ref{subsec:data_const}).

To approach the above problem, we describe our baseline models in terms of three components: a \textit{reader} module to read the text, a \textit{processor} module to perform reasoning over the text, and a \textit{classifier} module which takes a query entity tuple $(e_i, e_j)$ and predicts the relation $r_{(e_i, e_j)} \in R$. 

\subsection{LSTM baseline}

Firstly, we use a bidirectional LSTM (bi-LSTM) baseline \citep{hochreiter1997long}. 
We run the bi-LSTM \textit{reader} over the document $S$ to get the final document representation $h_t$ and the intermediate representations for each word $h_{w_i}$, which also contains the query entities $e_i$ and $e_j$. 
Next, we use a two layer MLP as the \textit{classifier} module which is provided the concatenation of the intermediate hidden states of the query entities as well as the final document representation: $p = \text{softmax}(\text{MLP}([\vec{h_t}, \vec{h_{e_i}}, \vec{h_{e_j}}]))$, where $\vec{h_t} = [\overrightarrow{h_t}, \overleftarrow{h_t}]$; $\vec{h_{e_k}} = [\overrightarrow{h_{e_k}}, \overleftarrow{h_{e_k}}]$ for $k \in (i,j)$.


\subsection{GNN Baseline}

To incorporate a relational inductive bias, we use a graph neural network (GNN) model as a second strong baseline. In this model, the \textit{reader} creates a unique edge representation from the relations by extracting the semantic information from the given text; the \textit{processor} is a GNN model that computes node representations, and the \textit{classifier} uses the node representations to classify the relation between the two query entities.

\xhdr{Reader Module}
For simplicity, we assume that every sentence in a story describes a relationship between two family members, and we use the words occurring between entities to extract a learned edge embedding. 
The edge embedding for each node pair is calculated by pooling over the set of words occurring between them in the text: $E_{i,j} = pool(\{w_1, w_2, ..., w_p\})$, where $pool$ is a differentiable pooling function (e.g., max-pooling or an attention-weighted average). 
After computing the $E_{i,j}$ embeddings from all sentences in the story, we obtain a graph $G$ with edge features given by $E_{i,j}$.

\xhdr{Processor Module}
The core component of the processor module is inherently a Message Passing Neural Network (MPNN) \citep{Gilmer2017}. The input to this processor is a set of node features $h = \{\vec{h}_1, \vec{h}_2, .. \vec{h}_K\}$, where $K$ is the maximum number of entities over the entire dataset. 
For each node, we  calculate the incoming message from each of its neighbors in the subset $\mathcal{N}(i)$ in the graph $G$ as:
\begin{equation}
    m_{i, j} = [E_{(i,j)}, \vec{p}_i, \vec{p}_j], \text{where} j \in \mathcal{N}(i),
\end{equation}
where $\vec{p_i}$ is an embedding indicating the position of the node $v_j$ within the graph $G$. The position information is represented as $\vec{p_i} = [\vec{e}_i, \vec{g}]$ where $\vec{e}_i$ is a random fixed embedding to represent the node, and $\vec{g}$ is a parameter we learn to represent the graph structure. Intuitively, the parameter $\vec{g}$ learns specific structural properties of the given graph and paired with a random embedding we can uniquely represent the node.

For each node, the incoming messages from its neighbors are averaged and combined together to form the final message: $M_i = \frac{1}{|\mathcal{N}(i)|} \sum_{j \in \mathcal{N}(i)} m_{i,j}$. We also experimented with an attention style message combination similar to Graph Attention Networks (GAT)\citep{Velickovic2017-mh}. Specifically, the individual attention scores are calculated as $a_{i,j} = W^{\top} tanh(W [[\vec{h}_i, \vec{p}_i];\vec{h}_j])$, and the final attention weight is calculated as a softmax over the neighbors: 
$\alpha_{i,j} = \frac{exp(a_{i,j})}{\sum_{k \in \mathcal{N}(i)} exp(a_{i,k})}
$. The combined message is a weighted sum over the attention: $M_i = \sum_{j \in \mathcal{N}(i)} m_{i, j} \alpha_{i,j}$

Finally, the node representation is updated based on an LSTM update function similar to \cite{Song2018-zs}:
\begin{equation}
    \vec{\hat{h}}_i = \text{LSTM}(M_i, (\vec{h}_i, c_i))
\end{equation}

\xhdr{Classifier module}
The above message passing and update function is performed over $r$ iterations, where $r$ is a hyperparameter. After the $r$ message-passing iterations,  the final representations of each node are averaged to get the final graph representation $H_T$. The final relation classifier takes a concatenation of $H_T$ and the node representations of the query entities: $p = \text{softmax}(\text{MLP}([H_T, H_{v_i}, H_{v_j}]))$.

\section{Experimental Results}
\begin{figure}[t]
    \centering
    \subfloat{{\includegraphics[width=0.45\textwidth]{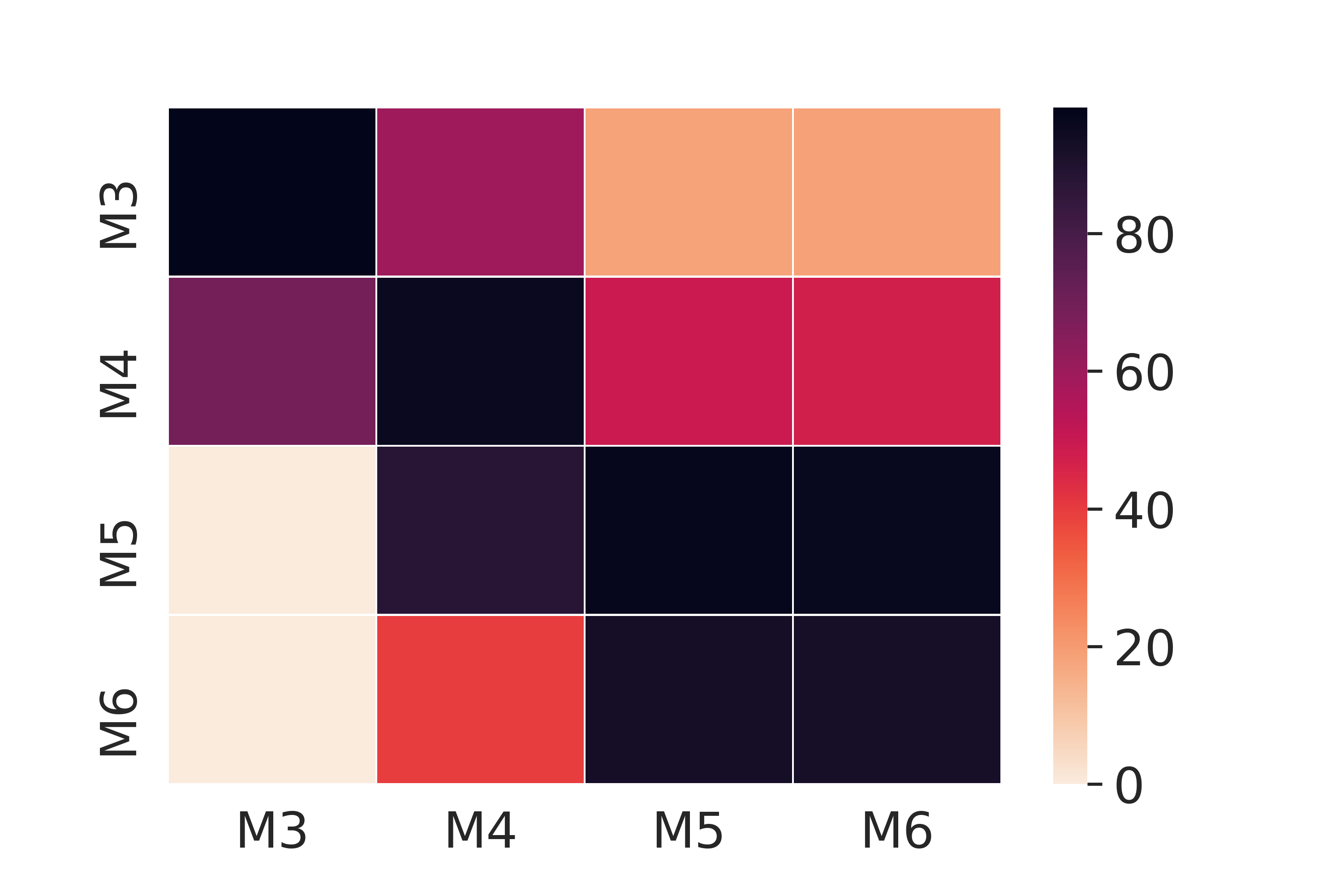} }}%
    \qquad
    \subfloat{{\includegraphics[width=0.45\textwidth]{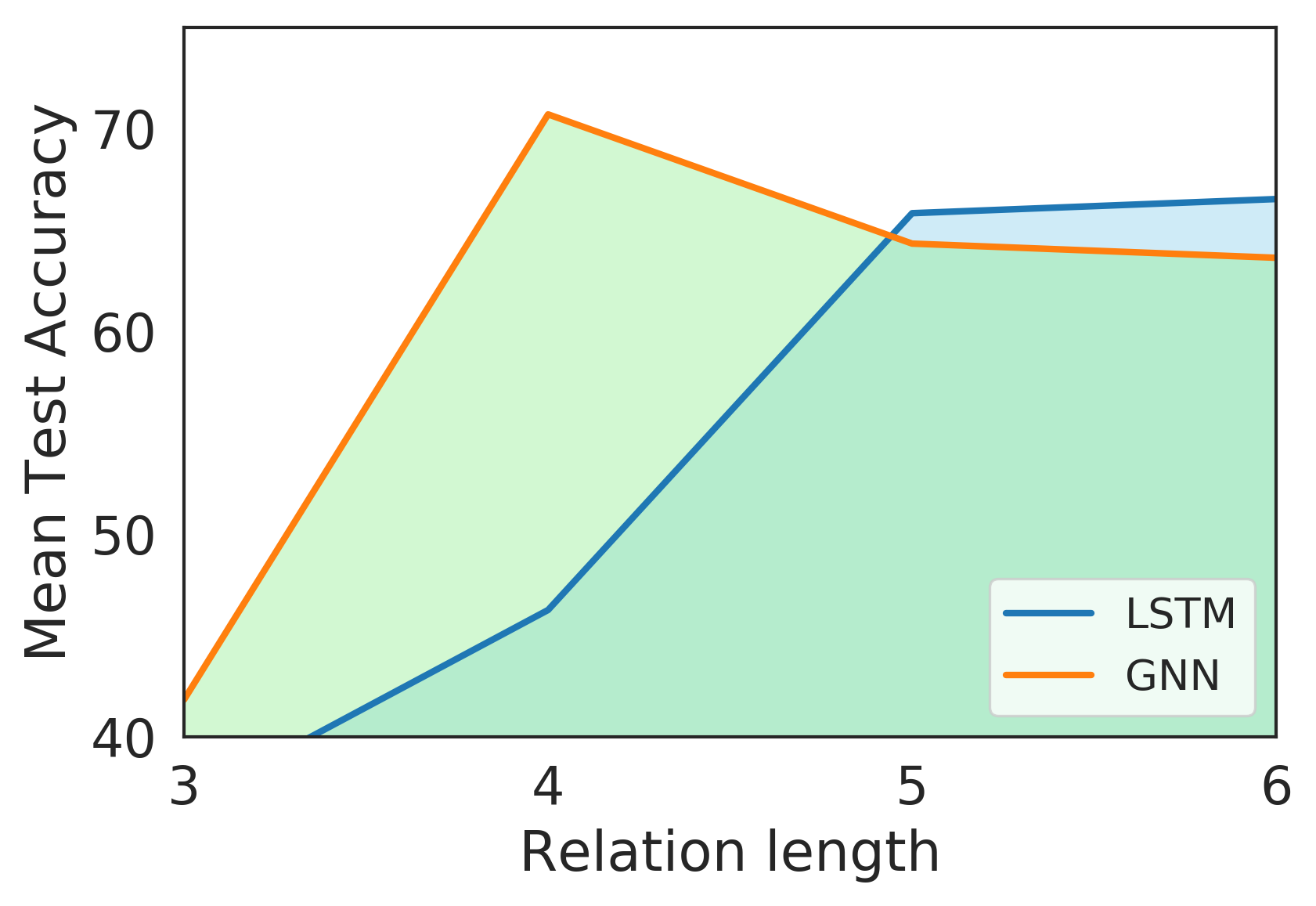} }}%
    \caption{{\bf Left:} the test accuracies for training data subset on the left axis and testing data subset on the right axis. {\bf Right:} the mean testing accuracy as an area plot accross the test data subsets.}%
    \label{fig:generalizability}%
    \vspace{-10pt}
\end{figure}

\begin{table}[t]
\caption{Generalizability experiment results}
\centering
\label{tab:results}
\resizebox{\textwidth}{!}{
\begin{tabular}{|l|llll|llll|llll|llll|}
\hline
Training      & \multicolumn{4}{c|}{$M_3$}                 & \multicolumn{4}{c|}{$M_4$}                                   & \multicolumn{4}{c|}{$M_5$}                                  & \multicolumn{4}{c|}{$M_6$}                                  \\ \hline
Testing       & $M_3$        & $M_4$       & $M_5$ & $M_6$ & $M_3$         & $M_4$        & $M_5$         & $M_6$         & $M_3$         & $M_4$         & $M_5$        & $M_6$        & $M_3$         & $M_4$         & $M_5$        & $M_6$        \\ \hline
LSTM          & \textbf{100} & 46.6        & 0     & 0     & 37            & \textbf{100} & 48            & 0             & 17.8          & 45.5          & \textbf{100} & \textbf{100} & 18.5          & 47.59         & \textbf{100} & \textbf{100} \\
GNN           & 98.3         & \textbf{69} & 0     & 0     & 59.8          & 95.3         & 87.8          & 39.9          & \textbf{18.4} & \textbf{49.4} & 96.8         & 92.7         & \textbf{18.5} & \textbf{47.6} & 96.2         & 92.2         \\
GNN-Attention & 99.8         & 67.79       & 0.001 & 0     & \textbf{65.7} & 98.2         & \textbf{89.9} & \textbf{44.1} & 16.8          & 47.1          & 99.1         & 97.9         & 17.2          & 46.29         & 98.7         & 97.2         \\ \hline
\end{tabular}
}
\end{table}



\xhdr{Setup}
In order to explicitly evaluate the models' performance on combinatorial generalization, we train on stories $M_k$ generated from relation paths of length $k$ and test on all the data subsets $M_{j}$ for $j \in k$, i.e., we test how well the models can generalize from stories bases on varying lengths of relation paths (see Section \ref{subsec:data_const} for further dataset details). 
 We used 100-dimensional word embeddings for both the models, which are randomly initialized and fine-tuned. We treat the entities as Cloze style placeholders so we invalidate the embedding for the entity words after each training iteration. For the LSTM baseline we choose a 2 layer Bi-LSTM with 50 hidden dimensions (i.e., the full intermediate hidden states are 100-dimensional). For the GNN baseline we set the individual node embeddings to have 100 dimensions. We set the position embedding $\vec{p}$ to have 15 dimensions where $\vec{e_i}$ is has 5 dimensions and $\vec{g}$ has 10 dimensions. The message passing is run for $r=6$ iterations. We train both models using Adam optimizer with a learning rate of 0.001.

\xhdr{Results}
From the results (Table \ref{tab:results}) we can see that the GNN baseline model outperforms the LSTM significantly in terms of generalizaion across data subsets. However, we also note that GNN model does not get perfect scores when tested on the same data subset. This indicates that the LSTM is extremely powerful at rote learning patterns, but fails to generalize. 
In contrast, the GNN baseline picks up the compositional elements to build a relation from smaller relation chains ($k \in (3,4)$) and generalizes it into larger relation chains ($k \in (5,6)$) with significantly higher accuracy. This is attributed to the inherent mechanism of relational reasoning with a GNN architecture.
We further illustrate the generalization capability of the GNN model in Figure \ref{fig:generalizability}.


\section{Conclusion}
We present a new dataset and highly extensible data generation approach to evaluate relational reasoning on language, providing a focused evaluation of a model's capacity for combinatorial generalization.
We also present two strong baseline algorithms and show that a model with a strong relational inductive bias achieves superior generalization performance.
As future work, we would like to extend our dataset and data generation approach by increasing the complexity of the relation set and adding crowdsourced (e.g., paraphrased) text for more natural language variation. 

\section*{Acknowledgement}

The authors would like to acknowledge Sanghyun Yoo, Byung In Yoo, Jehun Jeon and Young-Seok Kim from Samsung Advanced Institue of Technology (SAIT), Korea for their helpful comments, feedback and discussion. The first author would also like to thank the entire On-Device Language Learning team from SAIT, Korea led by Dr Young Sang Choi for hosting him in the Summer of 2018 at Suwon, Korea.

\bibliography{ref}
\bibliographystyle{unsrtnat}

\appendix
\section{Graph Generation}

As described in Section \ref{subsec:data_const}, we perform the graph generation over 3 steps: generation of random kinship graph, sampling of a simple path of $k$ edges, and replacing the edges with template-based natural language of the relation. The detailed pseudocode of data generation process is given in Algorithm \ref{alg:gen}. Specifically, we use two natural language template dictionaries: $TD$ for kinship relation dictionary and $AD$ for attribute relations. From $TD$ for each node pair and their relation from the set $R$ (refer Section \ref{subsec:data_const}), we sample a natural language description. For example, if $n_a$ and $n_b$ has a relation \textit{significant other}, then one possible sampled template would be ``A is the wife of B" depending on whether the gender of $n_a$ is female. $TD$ consists of several such templates in both gender relations for each $r \in R$.

\subsection{Distractor relations}

For each node, we sample $d$ distractor \textit{attributes}. These attributes can range from where the person works, which school they are alumni of, to political preferences (Republican / Democrat). We choose a set of 8 such distractor relations $D=$\{\textit{works\_at}, \textit{alumni\_of}, \textit{school}, \textit{location\_born}, \textit{preferred\_social\_media}, \textit{hobby}, \textit{sport}, \textit{political\_view}\}. The dictionary $AD$ contains templates for each of these distractor attributes which we sample as given in the Algorithm \ref{alg:gen}.

\begin{table}[h]
\caption{Total unique relation paths for $G = \{l_{max}=3, c_{min}=3, c_{max}=3\}$}
\centering
\label{tab:unique_paths}
\resizebox{0.7\textwidth}{!}{
\begin{tabular}{|c|c|c|}
\hline
\multicolumn{1}{|l|}{Path length} & \multicolumn{1}{l|}{Unique paths} & \multicolumn{1}{l|}{Unique paths with gender} \\ \hline
3                                 & 20                                & 40                                            \\
4                                 & 84                                & 168                                           \\
5                                 & 305                               & 610                                           \\
6                                 & 978                               & 1956                                          \\
7                                 & 2814                              & 5628                                          \\ \hline
\end{tabular}
}
\end{table}

\subsection{Relation combinations}

We sample relation path of length $k$ from the graph(s). Depending on the number of paths extracted from a graph, we generate more graphs if necessary to match the number of training-testing rows. Each row in the data, as described in $\textsc{Template Story Generation}$ procedure of Algorithm \ref{alg:gen}, consists of a pair of relation path of length $k$ and the target relation which is the relation among the first and the last node of the relation path. Each relation path does not contain duplicate nodes so as to remove cycles from the path. For example, a relation path of length 3 can have the following sequence: ${A \xrightarrow{child} B \xrightarrow{SO} C \xrightarrow{child} D}$ (where $SO$ denotes \textit{significant other}). In this example, the target relation would be $(A,D) \rightarrow grand[mother/father]$ depending on the gender of $A$. Note that in a target relation tuple we always refer the relation from the point of view of the first element. Thus if we inverse the above relation then the target changes : $(D,A) \rightarrow grand[son/daughter]$.

As we show in the example above, for each unique path we have two possible paths depending on the gender. This setup thus allows us to sample from the full possible combinations of kinship relations for each path length $k$. For example, total number of combinations of kinship relations for a graph of three levels and three children for each node is given in Table \ref{tab:unique_paths}.

\makeatletter
\def\BState{\State\hskip-\ALG@thistlm}
\makeatother

\begin{algorithm}
\caption{CLUTRR story generation process}
\label{alg:gen}
\begin{algorithmic}[1]
\Procedure{Graph Generation}{}
\State Initialize a graph $f$
\State Add a node $n$ to the graph $f$ which serves as head of the family
\State add the node $n$ in parents array $ps$
\For{$l = 0$ to $l_{max}$} \Comment{$l_{max}$ is the maximum number of levels, parameter}
\For{$p$ in  $[ps_{0}, ..., ps_{last}]$}:
\State Add a partner node $n_p$ of the opposite sex to the graph $f$
\State Add a \textit{significant other} edge between $p$ and $n_p$
\State Randomly determine the number of children $c_r$ of $(p,n_p)$ from a minimum $c_{min}$ and maximum $c_{max}$
\State Add $|c|$ number of child nodes with random gender in the graph $f$
\For{$c = 0$ to $c_{r}$ }: 
\State Add a \textit{child} edge between $(c, p)$
\State Add a \textit{child} edge between $(c, n_p)$
\EndFor
\State Replace parents array $ps$ with all children generated
\EndFor
\EndFor
\State return graph $f$
\EndProcedure
\Procedure{Path Sampling}{$f,k$} \Comment{k is the length of the relation we want to sample paths}
\State extract all siblings by grouping the nodes which have the same parent
\For{each node pair $(n_a, n_b)$ in the set of nodes $N$ of the graph $f$}
\If{there exists no edge between $n_a$ and $n_b$}
\If{there exists a shortest path between $n_a$ and $n_b$}
\State deduce the relation $r_{(n_a, n_b)} $ among the shortest path from the set of rules $R$
\State Add an edge between $(n_a, n_b)$ with this new relation
\EndIf
\EndIf
\EndFor
\State $P \leftarrow$ empty array
\For{each node pair $(n_a, n_b)$ in the set of nodes $N$ of the graph $f$}
\State Extract all simple paths $sp$ between the nodes $n_a$ and $n_b$
\State Sample all paths of length $k$
\State add the paths to $P$
\EndFor
Return all paths $P$
\EndProcedure
\Procedure{Template Story Generation}{$P$}
\State Initialize story-abstract pair $SP$ array
\For{each path $p_i \in P$}
\State Initialize empty string for story $S$
\State Initialize empty string for abstract $A$
\For{each sequential node pair $(n_a, n_b) \in p_i$}
\State Extract relation $r_{(n_a, n_b)}$ from the edge between the nodes $(n_a, n_b)$
\State For the relation and the gender of $n_b$, sample from the template dictionary $TD$ a placeholder sentence $t_p$
\State Replace the entity placeholders with the name of $r_{(n_a, n_b)}$ in $tp$
\State $S \leftarrow S + tp$
\State Sample $d$ attributes from each of the nodes $(n_a, n_b)$
\For{each attribute $d_i \in d$}
\State Sample a template from attribute dictionary $AD$ which is $td$. 
\State Replace the template with the node name and attribute name.
\State $S \leftarrow S + td$
\EndFor
\EndFor
\State Extract relation $r_{(n_{P[0]}, n_{P[-1]})}$ from the edge between the nodes $(n_{P[0]}, n_{P[-1]})$ which are the first and the last nodes of the path $P$
\State Similarly, sample from the template dictionary $TD$ a placeholder sentence $tp$ for the relation and gender of $n_{P[-1]}$.
\State $A = tp$
\State $SP \leftarrow SP + (S,A)$
\EndFor
Return $SP$
\EndProcedure
\end{algorithmic}
\end{algorithm}

\end{document}